\documentclass[a4paper,12pt]{article}
\usepackage[utf8x]{inputenc}
\usepackage{amsmath}
\usepackage{amsfonts}
\usepackage{url}

\begin{document}

\title{Using Automated Theorem Provers for Mistake Diagnosis in the Didactics of Mathematics}
\author{Merlin Carl}

\maketitle



\begin{abstract}
The Diproche system, an automated proof checker for natural language proofs specifically adapted to the context of exercises for beginner's students similar to the Naproche system by Koepke, Schr\"oder, Cramer and others, uses a modification of an automated theorem prover which uses common formal fallacies intead of sound deduction rules for mistake diagnosis. We briefly describe the concept of such an `Anti-ATP' and explain the basic techniques used in its implementation.
\end{abstract}

\section{Automated Proof Checking in the Didactics of Mathematics}

Learning how to prove is one major obstacle of the introductory phase of university education in mathematics. It requires practice, i.e. exercises, which need to be corrected, which is both an expensive and time-consuming task. This limits the way in which corrections can usually enter into the process of solving proof exercises as feedback. In the typical scenario, exercises are solved, handed in, corrected within a week and then given back to the students. In respondence to this problem, a number of automated tools for proof learning has been developed which provide instant feedback for a proposed solution, thus enabiling the student to have as many rounds of feedback and revision in a short period of time as he or she desires. We mention here the classical system GEOBEWEIS from the 90s \cite{Ho} for learning how to prove in elementary geometry, the more recent program QED-Tutrix \cite{QTHP} with the same purpose, Terence Tao's QED, which focuses on propositional and first-order logic \cite{QED}, the similarly logic-based system Edukera \cite{EHP}, the EProofs system \cite{EPHP}, \cite{NPKW} and the Concludio system currently under development in Germany, see \cite{Concludio}.
We emphasize at this point that such systems can only emulate human tutors to a very limited degree and should never be taken to replace them. Human tutors serve a number of functions which are extremely hard or downright impossible to transfer to a digital system, the ability to encourage students by serving as a role model for showing that what is asked of them is humany doable not being the least important among them.

Diproche is a further system in this spirit, which was first described in \cite{CK}. The main new feature of Diproche, which is implemented in SWI-Prolog, is the use of techniques from computational linguistics to allow for a free-text input in a restricted fragment of English specifically designed to capture formulations, formulas and figures of speech typically used in (template) solutions to beginner's exercises. The name, which is an acronym for `Didactical Proof Checking' is adpated from the name of the Naproche (`Natural Proof Checking') project, which is an ongoing cooperative project of mathematical logicians from the university of Bonn and linguists from Duisburg-Essen. The Naproche system can verify proofs in a controlled fragment of mathematical English and was mainly developed by Marcos Cramer in his dissertation \cite{Cr1}. Impressive as the recent examples of Naproche texts are, Naproche is, however, hardly usable for didactical purposes `as it is'. Some reasons for this are discussed elsewhere (see \cite{CK}); here, we only mention (1) the lack of control over acceptable proof steps due to the use of a professional (and strong) ATP in the background and (2) the lack of user feedback helpful for beginner's students, which is one of the points that we adress in this paper. Still, the main idea as well as the main architecture of Diproche are basically the same as for Naproche, and the internal representation format for proofs used by Diproche is also strongly inspired by the `Proof Representation Structures' used in Naproche. One should thus regard Diproche as a kind of didactical `offshoot' of Naproche.

Diproche provides several kinds of user feedback: A logical checking, which determines whether all proof steps are acceptable in the sense of the exercise, a type-checking to spot undeclared or wrongly used variables, a `goal-tracer' to determine whether the proof goal has been reached, automatically generated hints on proof strategies that could be applied when users get stuck, the possibility to obtain an intermediate step between a partial proof written up so far and the proof goal (obtained by using an ATP to automatically complete the proof and returning an intermediate step in the case of success) and a mistake diagnosis. The mistake diagnosis attempts to guess the misconception on the part of the user behind a certain non-verifiable proof step by using an anti-ATP, which is an automated theorem prover applying typical formal fallacies instead of sound proof rules. It is the goal of this paper to introduce the concept of an anti-ATP. To the best of our knowledge, Diproche is the first system (mis)using an ATP for mistake diagnosis in this way.\footnote{A pattern matching for diagnosing false algebraic manipulation rules is also used in the Concludio system, see \cite{Concludio}. 
	As a part of a friendly cooperation between the two projects, this idea was communicated to the Concludio developers by the author after it had been implemented in Diproche and was then used in Concludio.}

\section{Student-oriented feedback}

The mere fact that a proof step is labelled as false by is of little help to a student, especially in the beginner's phase, where the competence to diagnose for oneself what was wrong often yet has to be developed. Accordingly, a good tutor will often be able to reconstruct a student's motivation for a certain step and see the misconception behind it. This allows specifically adressing the misconception, both by further explanations and exercises. 

We give some examples of beginners student's mistakes that we frequently observed in our teaching experience:

\begin{enumerate}
	\item Suppose we have $(\phi\rightarrow\psi)$ and $\neg\phi$. Then we have $\neg\psi$.
	\item $(a+b)^{2}=a^2+b^2$
	\item $\neg(a\wedge b)$ is the same as $(\neg a \wedge \neg b)$.
\end{enumerate}

In each of these cases, anyone experienced in teaching mathematics will probably have a strong intuition of the underlying mistake: Namely, (1) comes from the misconception that implication `transports' falsity as well as truth and can nicely be encountered by any of a number of well-known everyday counterexamples (`If it rains, then the street is wet. If it does not rain, the street can still be wet for other reasons.'). (2) and (3) both seem to come from a general tendency to ignore semantics and apply supposed formal rules, in this case a `general distributivity' rule. (In particular, someone who makes these mistakes is probably more likely to also believe that the derivation of $f\cdot g$ is $f^{\prime}\cdot g^{\prime}$ etc.) It can again be adressed by giving counterexamples (a numerical one for (2) and again some everyday situation for (3)), but this should ideally be supplemented by emphasizing that the occuring symbols have a meaning and that one can (and should) try to understand the meaning of a formula. In the case of (2), one could, e.g., remind that $(a+b)^2=(a+b)(a+b)$ and further illustrate this by drawing a square side length $(a+b)$ and marking the subsquares of dimensions $a\times a$ and $b\times b$, thus making clear that this does not exhaust the full figure. Moreover, one can offer and dicsuss a number of related situations triggering the same mistake in order to make the student aware of the problem and sensible to it. (For example, one could consider the questions whether $\sqrt{a+b}=\sqrt{a}+\sqrt{b}$, $a\frac{b}{c}=\frac{ab}{ac}$, $\int_{0}^{1}f(x)g(x)dx=\int_{0}^{1}f(x)dx\int_{0}^{1}g(x)dx$, $2^{m+n}=2^{m}+2^{n}$ or whether $\neg(A\rightarrow B)$ is the same as $\neg A\rightarrow \neg B$.)

For an automatic proof tutoring system like Diproche, it is desirable to at least partially emulate this ability of human tutors automatically. 

\section{Automatization of Mistake Diagnosis}

An automated theorem prover (ATP) applies inference rules to axioms to deduce new formulas. In didactical proof checking, a controlled ATP is used to verify whether a step claimed by the user does indeed follow from the assumptions available at this point. To this end, the ATP is taylor-made to accept exactly those steps that can count as `elementary' at a certain stage of education. Since the set of these steps develops rapidly during mathematical education (for example, it might make sense to give something like the de Morgan rule as an exercise at a very early stage, while later on, one should be able to use it freely and without further mentioning), possibly even from one exercise to the next, one can additionally specify for each exercise the set of admissible deduction rules. Similarly then, the mistake diagnosis of Diproche uses an `ATP' which, instead of correct inference rules, works with common formal fallacies. We split the description into two parts, one for logical fallacies and one for false algebraic and numerical manipulations. 

Within Diproche, the anti-ATP then works like this: When the ATP fails to verify a certain proof step, the mistake diagnosis is started. Thus, the anti-ATP is applied to the same step. When the anti-ATP succeeds in `verifying' the deduction step in question by one of its rules, the internal index of that rule is returned and a message for the user explaining the type of the suggested fallacy is written on the screen.

\subsection{Diagnosis of logical fallacies: The ``Anti-ATP''}

Like a sound inference rule, a formal fallacy is a formal rule. As mentioned above, a mistake diagnosis is technically realized in Diproche by using an ATP with formally fallacious rules.	

We give here a list of sample rules that are included in the current version of the Anti-ATP:

\begin{itemize}
	\item (Inverse Contraposition)
	\begin{tabular}{l}
		$A\rightarrow B$\\
		\hline \\
		$\neg{A}\rightarrow\neg{B}$	
	\end{tabular} 
\quad 
 \begin{tabular}{l l}
	$\neg A$ & $(A->B)$ \\
	\hline \\
	$\neg B$ & \\
\end{tabular}

\item (Inverse Implication)
\begin{tabular}{l}
	$A\rightarrow B$\\
	\hline \\
	$B\rightarrow A$
\end{tabular}

\item (Exclusive Reading of `or')
\begin{tabular}{l l}
	$A$ & $A\vee B$ \\
	\hline\\
	$\neg{B}$
\end{tabular}

\item (Misinterpretation of Implication)
\begin{tabular}{l}
	$\neg{A}$ \\
	\hline \\
	$\neg(A\rightarrow B)$ \\
\end{tabular}

\item (False Distributivity of Negation)
\begin{tabular}{l}
$\neg(A\wedge B)$ \\
\hline \\
$\neg{A}\wedge\neg{B}$	
\end{tabular}

\item (Quantifier Switch)
\begin{tabular}{l}
	$\forall{x}\exists{y}\phi$ \\
	\hline \\
	$\exists{y}\forall{x}\phi$
\end{tabular}

\item (False Quantifier Negation)
\begin{tabular}{l}
	$\neg\forall{x}\phi$\\
	\hline \\
	$\forall{x}\neg\phi$
\end{tabular}
\quad
\begin{tabular}{l}
	$\neg\exists{x}\phi$\\
	\hline \\
	$\exists{x}\neg\phi$
\end{tabular}

\item (Confusion of Subset and Element Relation)
\begin{tabular}{l}
	$A\subseteq B$ \\
	\hline \\
	$A\in B$
\end{tabular}
\quad 
\begin{tabular}{l}
	$A\in B$\\
	\hline \\
	$A\subseteq B$
\end{tabular}

\item (Transitive Usage of Element Relation)
\begin{tabular}{l l}
	$A\in B$ & $B\in C$ \\
	\hline \\
	$A\in C$
\end{tabular}

\end{itemize}

\subsection{Diagnosis of false algebraic and numerical manipulations}

Besides fallacious logical inferences, a common type of mistake among beginner's students are false algebraic manipulations. Quite frequently, these arise out of an application of a systematically false manipulation rule. In this sense, they are similar to the formal fallacies processed by the anti-ATP. Such mistakes can be diagnosed by a submodule of the anti-ATP, which currently goes by the name `antiterms'. A new feature is that a false algebraic manipulation should still be recognizable when combined with correct manipulation steps, as in $\frac{1}{2}+\frac{1}{2}=\frac{2}{4}$. For this reason, a submodule for the verification of term manipulations is used when attempting to match a certain manipulation step with a false manipulation rule. Thus, $\frac{1}{2}+\frac{1}{2}=\frac{1+1}{2+2}$, $\frac{1}{2}+\frac{1}{2}=\frac{2}{2+2}$, 
$\frac{1}{2}+\frac{1}{2}=\frac{2}{4}$ and even
$\frac{1}{2}+\frac{1}{2}=\frac{1^{2}+1^{3}}{\sqrt{4}+(3-1)}$ will be matched with the `component-wise addition of fractions'-rule.

We again list some sample rules included in the current version.\footnote{This list is by no means comprehensive.}

\begin{itemize}
	\item (Component-wise Addition of Fractions)
	$\frac{a}{b}+\frac{c}{d}=\frac{a+c}{b+d}$
	\item (Additive Cancellation)
	$\frac{a+b}{b+d}=\frac{a}{b}$ 
	\item (Base Cancellation)
	$\frac{a^{m}}{a^{n}}=\frac{m}{n}$
	\item (Exponent Cancellation) $\frac{a^{mn}}{a^{m}}=a^{n}$
	\item (Distributive Use of Exponentiation) $(a+b)^{n}=(a^{n}+b^{n})$.
	\item (Distributive Use of Multiplication over Exponentiation) 
	$a\cdot\frac{b}{c}=\frac{a\cdot b}{a\cdot c}$
\end{itemize}


By a recursion on the construction of formulas, the antiterms module is recursively applied to parts of formulas; thus, when a false manipulation rule is applied merely to a certain proper subterm, the mistake can still be spotted. For example, $(3(\frac{1}{2}+\frac1{1}{2}))^{2}=(3\cdot\frac{2}{4})^{2}$ would still be recognized as an instance of `component-wise addition of fractions.
However, the simultaneous application of several fallacious rules to different part of a term is currently not recognized; the reason for this is that we suspect such kinds of diagnosis obtained by several applications the antiterms module at the same time to be too hypothetical, which diminuishes the reliability of the feedback (see the discussion below).

For example, $5\cdot (\frac{1}{2}+\frac{1}{2})^2=5\cdot\frac{2}{4}^2$ would be recognized as an instance of adding fractions component-wise. Currently, the diagnosis will only work when only one subterm is falsely manipulated. It would of course be easy to diagnose several mismanipulations at the same time, but this would become too speculative, in particular given that, in many cases, several false rules may explain a mistake (``you might have combined ... with ... and ...'').

\subsubsection{Diagnosing mistakes by types}

Many mistake patterns can be subsumed under more general types. These types are automatically recognizable as well and can also be used to diagnose mistakes in domains to which the anti-ATP was not specifically set up.

We again mention some examples.

\bigskip
\textbf{General distributivity}

\begin{itemize}
	\item $(a+b)^2=(a^2+b^2)$
	\item $(fg)^{\prime}=f^{\prime}g^{\prime}$
	\item $\sqrt{a+b}=\sqrt{a}+\sqrt{b}$
	\item $\frac{a}{b}+\frac{c}{d}=\frac{a+c}{b+d}$
	\item $\neg(a\wedge b)$ is equivalent to $(\neg a\wedge \neg b)$
\end{itemize}


\bigskip
\textbf{General commutativity}

For example, the composition of functions, matrix multiplication or group or ring operations are taken to be commutative or used as if they were.

\bigskip
\textbf{General monotonicity}

Examples would be to deduce that, if $b>c$, then 

\begin{itemize}
	\item $a-b>a-c$ 
	\item $a^{b}>a^{c}$ 
	\item $\frac{a}{b}>\frac{a}{c}$
\end{itemize}

The automatic recognition of such types of mistakes can be realized by an easy adaptation of the antiterms module. For example, a Prolog clause for recognizing a use of the general distributivity rule looks like this:

\begin{center}
false\_manipulation(gen\_distr,[A,Op0,[B,Op1,C]],[[A,Op0,B],Op1,[A,Op0,C]]):-\\
                  binary\_operator(Op0),
                  binary\_operator(Op1).
\end{center}

Here, gen\_distr is the index of the manipulation rule, which is applied to equalities of the form $[T,=,S]$ by running false\_manipulation(X,T,S).

Though it is not hard to implement such rules (as one can see from this example), they are not used in the current Diproche version, as it does not seem to be an easy task to generate a user feedback based on them that is likely to be helpful for a beginner student that makes such mistakes.

\subsubsection{Diagnosing by false analogy}

We mention here a possibility to take the idea of the last section a bit further.
As a mistake, $(a+b)^2=(a^2+b^2)$ can also be explained as a false analogy with $(a+b)\cdot 2=(a\cdot 2)+(b\cdot 2)$. Thus, a law that holds for addition has been taken to also hold true for exponentiation.

Such `analogy mistakes' can also be automatically recognized.

The Prolog clause for the diagnosis of false analogies looks like this:

\bigskip

false\_manipulation\_by\_analogy([false\_analogy,Op0,Op1],Term0,Term1,Anz,Vss):-

member(Op0,[exp,-,+,/,*]),

member(Op1,[+,exp,*,/,-]),

\textbackslash+ Op0=Op1,

replace\_operators(Term0,[[Op0,Op1]],Term00),

replace\_operators(Term1,[[Op0,Op1]],Term11),

check\_inequality\_chain([Term00,=,Term11],Anz,Vss).

\bigskip

Thus, the mistake diagnosis via false analogy works like this: If $T_{1}$ and $T_{2}$ are terms and the claim $T_{1}=T_{2}$ is not verified by the ATP, it is tried whether a correct equality can be obtained by replacing some arithmetical  operation Op$_{0}$ in $T_{1}$ and $T_{2}$ by another such operation Op$_{1}$. When this is indeed possible, the message is returned that $T_{1}=T_{2}$ can be obtained from a correct equality by confusing Op$_{0}$ with Op$_{1}$.

In the current implementation, only one operator is replaced. While it might be conceivable that there are conceptual mistakes that come from a simultaneous confusion of several operators - for example, a claim like $a+(b\cdot c)=(a\cdot b)+(a\cdot c)$ could plausibly be explained by false analogy with $a\cdot (b+c)=(a\cdot b+(a\cdot c)$, where $+$ and $\cdot$ are interchanged - this seems again to introduce a level of speculativeness that decreases the probability of a cognitively accurate diagnosis - i.e. one that actually represents the misconception behind the mistake - to an extent that this seems of little value unless further data about the mistake patterns of a certain user are available. 

We note here that this problem even arises for only one replaced operator: For example, $(a+b)^2=(a^2+b^2)$ can be explained both by false analogy with $(a+b)\cdot 2=(a\cdot 2)+(b\cdot 2)$ and $(a\cdot b)^2=(a^2\cdot b^2)$, although only the former option seems to occur frequently.


\section{Discussion}

Like human correctors, there are several ways in which the anti-ATP can err:
 For example, at lest in principle, it might happen that the anti-ATP wrongly diagnoses a correct inference step as mistaken, namely when the ATP fails to recognize the step as correct and the step can \textbf{also} be obtained by a common formal fallacy or incorrect manipulation. For example, if the ATP is to weak to recognize that $(a+b)^2=a^2+b^2$ (mod $2$) holds true, this would be marked as a distributive application of exponentiation. Also, there could be several fallacies that explain a certain mistake. Consequently, it is still up to the user to make use of the feedback and see whether the proposed formal fallacy was indeed the reason for the mistake (or to provide more intermediate steps if there was no mistake, e.g., by writing $(a+b)^2=(a^2+2ab+b^2)=(a^2+b^2)$ (mod $2$)).

\section{Further work}

The fallacies covered by the anti-ATP in its current form were obtained by the personal experience of the author with correcting exercises and exams as well as occasional hints by his colleagues. It would certainly be preferable to systematically amend and back up the selection. Unfortunately, little information seems to be available on formal deductive fallacies that typically occur in exercises. To this end, an empirical study is currently planned.

In contrast, false numerical manipulations have received a lot of attention, see, e.g., \cite{PW}. It will be the content of future work to integrate the empirical work in didactics into the antiterms module.


A further potential extension would be to allow for a mistake diagnosis also in the case of a combination of correct deductions with fallacies or of several false manipulations. The risk, as already mentioned above, is that this is too speculative to be helpful. Human tutors may be able to pose reliable diagnoses when they possess a lot of experience, both in general and with the particular student. In principle, one could try to emulate this by monitoring the use of the system and applying tools from machine learning to improve the diagnosis. However, whether or not such an approach will yield the desired results has to be determined by experience.

Finally, it might even be possible to automatically extend the set of recognizable fallacies by building up a database of attempted false steps and searching for reoccuring formal patterns. How well this works out in practice will also need to be determined in future work.













\begin{thebibliography}{}
           \bibitem[Br]{Br} G. Brandl. Analyse von Rechenfehlern im Grundschulbereich. Ein Beitrag zur Behebung von Rechenschw\"ache/Arithmastenie. Ars Una (1992)
	\bibitem[CK]{CK} M. Carl, R. Krapf. Das Diproche-System – ein automatisierter Tutor für den Einstieg ins Beweisen. submitted, (2019)
           \bibitem[Concludio]{Concludio} Concludio Homepage. \url{https://www.concludio.education/}
	\bibitem[Cr1]{Cr1} M. Cramer. Proof-checking mathematical texts in controlled natural language. PhD thesis (2013)
\bibitem[CFKKSV]{CFKKSV} M. Cramer, B. Fisseni, P. Koepke, D. Kühlwein, B. Schröder and J. Veldman.  The Naproche Project – Controlled Natural Language Proof Checking of Mathematical Texts. Proceedings of the Controlled Natural Language (CNL) Workshop. (2009)
           \bibitem[EPHP]{EPHP} E-Proofs Homepage. \url{http://e-proof.weebly.com/}
          \bibitem[NPKW]{NPKW}  E. Niehaus, M.Platz, M. Krieger, K. Winter. Elektronische Beweise in der Lehre. Beitr\"age zum Mathematikunterricht. (2016)
	\bibitem[QED]{QED} T. Tao. QED Homepage. \url{https://www.math.ucla.edu/~tao/QED/QED.html}
	\bibitem[QTHP]{QTHP} QED-Tutrix Homepage. \url{http://turing.scedu.umontreal.ca/qedx/}
	\bibitem[EHP]{EHP} Edukera Homepage. \url{https://www.edukera.com/}
	\bibitem[Ho]{Ho} G. Holland. GEOLOG-WIN : Konstruieren, Berechnen, Beweisen, Problemlösen mit dem Computer im Geometrie-Unterricht der Sekundarstufe. D\"ummler-Verlag. (1996)
	\bibitem[CHK]{CHK} F. Grewing. Concludio Homepage. \url{https://www.concludio.education/}
	\bibitem[PW]{PW} F. Padberg, S. Wartha. Didaktik der Bruchrechnung. Springer Spektrum (2017)

\end{thebibliography}
\end{document}